\documentclass{article}

\usepackage[preprint]{neurips_2026}

\usepackage[utf8]{inputenc} 
\usepackage[T1]{fontenc}    
\usepackage{hyperref}       
\usepackage{url}            
\usepackage{booktabs}       
\usepackage{amsfonts}       
\usepackage{amsmath}        
\usepackage{nicefrac}       
\usepackage{microtype}      
\usepackage{xcolor}         
\usepackage{enumitem}       
\usepackage{multirow}       
\usepackage{colortbl}       
\usepackage{graphicx}       
\usepackage{titlesec}       

\titlespacing*{\section}{0pt}{2ex plus 0.5ex minus 0.3ex}{1ex plus 0.2ex}
\titlespacing*{\subsection}{0pt}{1.5ex plus 0.4ex minus 0.2ex}{0.75ex plus 0.2ex}

\title{Learning the Signature of Memorization in Autoregressive Language Models}

\author{
  David Ilić\textsuperscript{1} \quad
  Kostadin Cvejoski\textsuperscript{1} \quad
  David Stanojević\textsuperscript{1} \quad
  Evgeny Grigorenko\textsuperscript{1} \\[1ex]
  \textsuperscript{1}JetBrains Research \\
  \texttt{\{david.ilic, kostadin.cvejoski, david.stanojevic, evgeny.grigorenko\}@jetbrains.com}
}

\begin{document}

\maketitle

\begin{abstract}

All prior membership inference attacks for fine-tuned language models use hand-crafted heuristics (e.g., loss thresholding, Min-K\%, reference calibration), each bounded by the designer's intuition. We introduce the first transferable learned attack, enabled by the observation that fine-tuning any model on any corpus yields unlimited labeled data, since membership is known by construction. This removes the shadow model bottleneck and brings membership inference into the deep learning era: learning what matters rather than designing it, with generalization through training diversity and scale.
We discover that fine-tuning language models produces an invariant signature of memorization detectable across architectural families and data domains. We train a membership inference classifier exclusively on transformer-based models. It transfers zero-shot to Mamba (state-space), RWKV-4 (linear attention), and RecurrentGemma (gated recurrence), achieving 0.963, 0.972, and 0.936 AUC respectively. Each evaluation combines an architecture and dataset never seen during training, yet all three exceed performance on held-out transformers (0.908 AUC). These four families share no computational mechanisms, their only commonality is gradient descent on cross-entropy loss. Even simple likelihood-based methods exhibit strong transfer, confirming the signature exists independently of the detection method.
Our method, Learned Transfer MIA (LT-MIA), captures this signal most effectively by reframing membership inference as sequence classification over per-token distributional statistics. On transformers, LT-MIA achieves 2.8$\times$ higher true positive rate at 0.1\% false positive rate than the strongest baseline. The method also transfers to code (0.865 AUC) despite training only on natural language texts. Our results imply that leakage from memorization is intrinsic to cross-entropy training; architectural innovation within this paradigm did not escape our attack. 
Code and trained classifier available at \href{https://github.com/JetBrains-Research/learned-mia}{this URL}.

\end{abstract}

\section{Introduction}

Membership inference attacks (MIAs) determine whether a specific text appeared in a model's training data, with applications in privacy auditing, copyright detection, and data governance. All existing MIA methods for language models are hand-crafted heuristics: loss thresholding~\citep{yeom2018privacy}, Min-K\%~\citep{shi2024detecting}, Zlib normalization~\citep{carlini2021extracting}, reference-based calibration~\citep{carlini2022membership}, and EZ-MIA~\citep{ilic2026ezmia}. Each designs a statistic by intuition and thresholds it, meaning that performance is bounded by the designer's insight into what signals membership. Learned attacks that train classifiers on shadow models with known membership outperform heuristics ~\citep{shokri2017membership, salem2019mlleaks}, but require training many models with known membership, which is prohibitive at LLM scale. The implicit assumption has been that learned MIA requires shadow models, and therefore cannot scale to language models.

We introduce the first transferable learned membership inference attack for fine-tuned language models. The key insight is that MIA has unlimited labeled training data by construction: fine-tuning any model on any corpus yields perfect membership labels. This removes the scalability bottleneck faced by shadow model approaches and enables standard deep learning benefits: learning what matters rather than designing it, generalization through training diversity and scale, and transfer to settings never seen during training.

Our membership inference attack, Learned Transfer MIA (LT-MIA), trained exclusively on transformer language models, transfers zero-shot to architectures it has never seen, including state-space models with no attention mechanism. LT-MIA achieves 0.963 area under the ROC curve (AUC) on Mamba-2.8B~\citep{gu2023mamba}, 0.972 on RWKV-4-3B~\citep{peng2023rwkv}, and 0.936 on RecurrentGemma-2B~\citep{botev2024recurrentgemma} when detecting whether a text was used during fine-tuning, surprisingly exceeding performance on held-out transformers (0.908 AUC). This transfer is not an artifact of our method's design: even simple loss thresholding (Loss) achieves 0.867 AUC on Mamba, and reference-based loss calibration (RefLoss) reaches 0.892. 

The four architectural families share no computational mechanisms. Transformers compute attention through pairwise token interactions with quadratic complexity~\citep{vaswani2017attention}. Mamba uses selective state spaces that evolve hidden states through input-dependent dynamics, with no attention mechanism whatsoever~\citep{gu2023mamba}. RWKV replaces softmax attention with linear recurrence, enabling constant-memory inference~\citep{peng2023rwkv}. RecurrentGemma combines gated linear recurrence with local attention windows~\citep{botev2024recurrentgemma}. These models differ in how they process context, how they scale with sequence length, and how information flows through layers. Their only commonality is the training objective: gradient descent on cross-entropy loss.

The implication is that memorization produces a signature in the output distribution, not the computation. Cross-entropy training induces a distributional shift toward training data: any model trained this way will assign higher likelihood to training samples relative to its pre-trained state, and this shift is detectable regardless of how the model computes that likelihood. Feature ablation confirms that comparison features (those relating target to reference model outputs) dominate across all four families, with the same feature importance hierarchy appearing in transformers, state-space models, linear attention, and gated recurrence alike.

\begin{figure}[t]
    \centering
    \includegraphics[width=\linewidth]{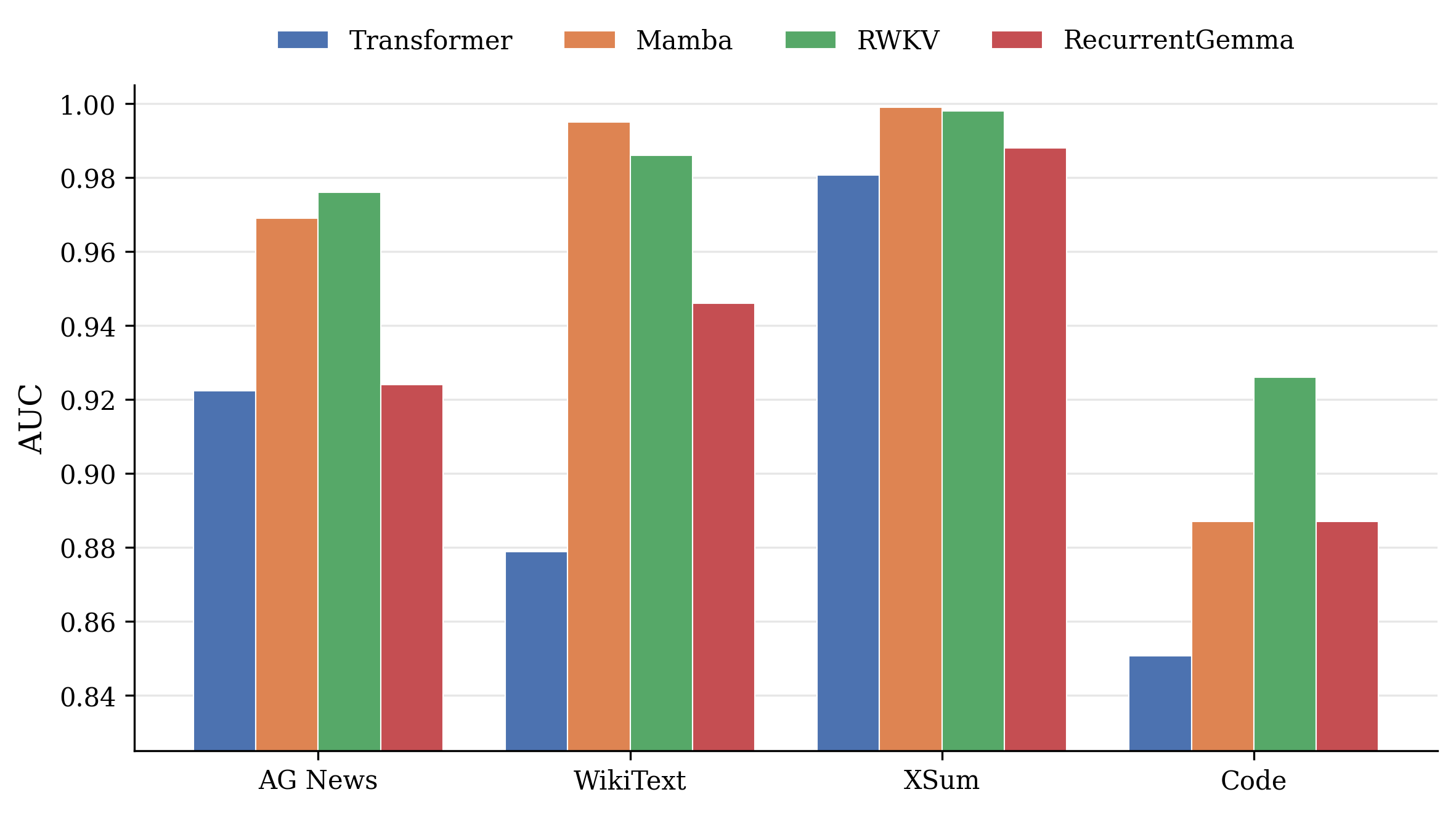}
    \caption{LT-MIA transfers from transformers to non-transformer architectures, exceeding transformer performance on every text dataset. Blue bars show held-out transformer results (mean over 7 models), while the remaining bars show zero-shot transfer to Mamba-2.8B (state-space), RWKV-4-3B (linear attention), and RecurrentGemma-2B (gated recurrence). The classifier was trained exclusively on transformers and textual data.}
    \label{fig:hero}
\end{figure}

To train our classifier, we reframe membership inference as sequence classification over per-token features. Given a text and black-box access to a fine-tuned target model and its pre-trained reference, we extract features at each token position capturing how the two models' predictions diverge. A lightweight transformer classifier processes this feature sequence, learning to weight positions adaptively rather than aggregating to a single statistic. Sequence modeling contributes 5.0 AUC percentage points over mean pooling (Section~\ref{sec:analysis}), with the transformer's advantage likely stemming from its adaptive weighting ability.

Training diversity proves essential. We train on 30 model-dataset combinations with 600 samples each, which substantially outperforms training on 1 combination with 18,000 samples (0.860 vs.\ 0.796 AUC on held-out combinations), despite identical data volume. Diversity filters model-specific artifacts (tokenizer idiosyncrasies, architectural biases in logit scales), retaining only the memorization signal shared across models. This filtering may explain why transfer succeeds: by learning to ignore transformer-specific patterns during training, the classifier relies on the invariant likelihood shift that exists in any model trained on cross-entropy loss.

The signal itself exists across architectures, data domains, and attacks, but LT-MIA captures it most effectively. On held-out transformer combinations, LT-MIA achieves 0.908 mean AUC, outperforming the strongest baseline (EZ-MIA) by 2.1 points. The gap widens at stricter operating points: 1.6$\times$ higher TPR at 1\% FPR and 2.8$\times$ at 0.1\% FPR. On non-transformer architectures, performance is stronger still, with all three families exceeding the transformer mean. The method also transfers to code (0.865 AUC) despite training only on natural language text.

We make four contributions. First, we identify the key enabler: fine-tuning provides unlimited labeled training data, removing the shadow model bottleneck that blocked learned approaches. Second, we demonstrate that memorization produces an architecture-invariant signature: a detector trained on transformers transfers zero-shot to state-space models, linear attention, and gated recurrence, with all tested likelihood-based methods exhibiting this transfer. Third, we move membership inference for language models from hand-crafted heuristics into the deep learning era, introducing Learned Transfer MIA (LT-MIA), the first transferrable learned attack. Fourth, we demonstrate that training diversity, not pure scale, enables generalization: 30 combinations with 600 samples each outperforms 1 combination with 18,000 samples, and show that sequence classification over per-token features outperforms aggregation-based methods, contributing 5.0 AUC percentage points. Our results imply that privacy risk from memorization is intrinsic to cross-entropy training, implying that architectural innovation within this paradigm did not escape membership inference.

\section{Related Work}

\textit{Membership inference foundations.} \citet{shokri2017membership} introduced the shadow model paradigm: train shadow models with known membership, then train a classifier on their outputs. This enabled learned attacks but required training many shadow models, which is prohibitive at LLM scale. \citet{yeom2018privacy} connected MIA success to overfitting, proposing simple loss-threshold attacks that require no training. \citet{feldman2020does} showed that memorization is \emph{necessary} for near-optimal generalization on long-tailed distributions, a property of the learning problem, not model capacity. If memorization arises from the objective, its signature should transfer across architectures sharing that objective.

\textit{Hand-crafted MIA for language models.} Subsequent work developed increasingly sophisticated heuristics requiring no attack training. Reference-based calibration compares target model outputs to a reference: zlib entropy~\citep{carlini2021extracting}, Min-K\% Prob~\citep{shi2024detecting}, vocabulary-level normalization~\citep{zhang2024mink}, and window-based comparison~\citep{chen2026wbc}. EZ-MIA~\citep{ilic2026ezmia} aggregates probability shifts at positions where the target model errs. These methods apply to any model but share a fundamental limitation: each is a hand-crafted statistic. Performance is bounded by the designer's intuition about what signals membership. \citet{duan2024membership} showed existing methods achieve near-random performance on pretraining data at scale.

\textit{The scalability gap.} Learned attacks outperform hand-crafted heuristics in other domains~\citep{salem2019mlleaks}, but have not scaled to LLMs. Shadow model training is prohibitively expensive, and classifiers trained on one model-dataset combination overfit to that setting. Prior transfer work assumed architectural similarity between shadow and target models~\citep{shokri2017membership}. Transfer across fundamentally different architectures has not been demonstrated. We address this gap by reframing MIA as sequence classification over per-token features, using deep models trained on diverse combinations to filter architecture-specific artifacts.

\section{Method}

\begin{figure}[t]
    \centering
    \includegraphics[width=\linewidth]{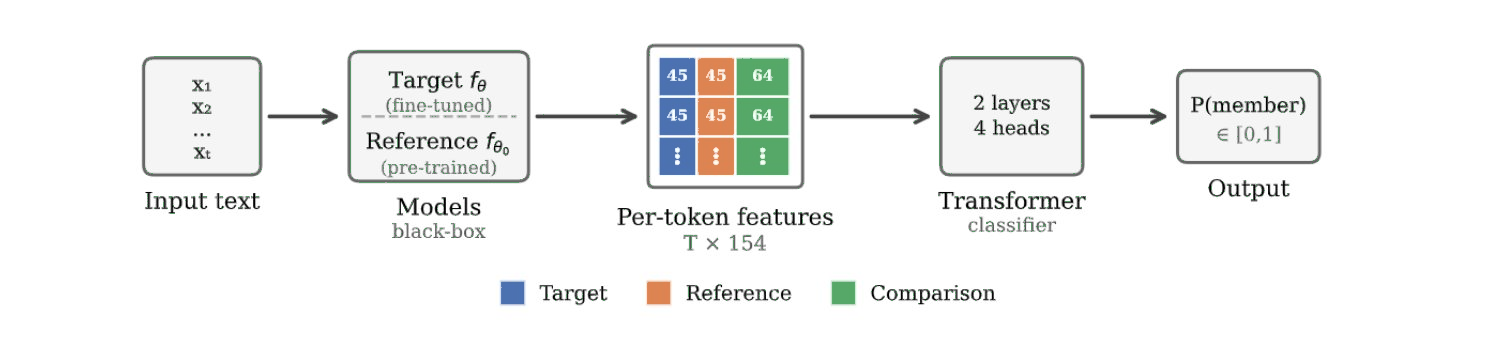}
    \caption{LT-MIA pipeline. Given a text sample and black-box access to a fine-tuned target model and its pre-trained reference, we extract 154-dimensional feature vectors at each token position capturing how the two models' predictions diverge. A lightweight transformer encoder processes this feature sequence to predict membership.}
    \label{fig:pipeline}
\end{figure}

Figure~\ref{fig:pipeline} illustrates the LT-MIA pipeline. Given a text sample and black-box access to both a fine-tuned target model and its pre-trained reference, we extract a feature vector at each token position and classify the resulting sequence.

\subsection{Threat Model}
\label{sec:threat_model}

We consider an adversary performing fine-tuning membership inference: given a candidate text $x$, determine whether $x$ was included in the fine-tuning dataset $\mathcal{D}_{\text{ft}}$.

\textit{Adversary capabilities.} The adversary has black-box query access to the fine-tuned target model $f_\theta$ and a pre-trained reference model $f_{\theta_0}$ sharing the same architecture. Both models return full vocabulary logits at each token position. The adversary can query them with arbitrary inputs.

\textit{Adversary knowledge.} The adversary knows the model's architecture family (to identify the appropriate reference checkpoint) and possesses candidate texts to query, for instance, a copyright holder checking whether their content was used for fine-tuning.

\textit{What is not required.} The adversary does not need white-box access, knowledge of the training data distribution, or the ability to train shadow models of the target.

\textit{Train-once-attack-anywhere.} Unlike shadow model approaches, our threat model decouples attack training from target access. The adversary trains a classifier once on diverse public models, then deploys against any target, including architectures never seen during training, without retraining. This is the capability our experiments validate.

\subsection{Per-Token Feature Extraction}
\label{sec:features}

Let $f_\theta$ denote the fine-tuned target model and $f_{\theta_0}$ the pre-trained reference sharing the same architecture. For input sequence $x = (x_1, \ldots, x_T)$, we extract 154 features at each position $t$, capturing how the two models' predictions diverge.

The features, normalized for scale invariance, include: 
\begin{itemize}
    \item per-token losses from both models and their difference
    \item global statistics (mean, standard deviation of each loss term, total log-likelihood ratio) broadcast to every position
    \item the logit and rank of the ground-truth token $x_t$ in both distributions
    \item top-20 and bottom-20 logits from each model
    \item cross-model ranks: for each model's top-20 and bottom-20 tokens, their rank in the other model's distribution
\end{itemize}

\subsection{Architecture \& Training}

The feature sequence is projected to dimension $d = 128$, combined with sinusoidal positional encodings~\citep{vaswani2017attention}, and processed by a 2-layer transformer encoder with 4 attention heads and feedforward dimension 256. A learned query computes attention over encoder outputs, that is then fed as a weighted sum into a two-layer MLP classifier. In total, the attack model has fewer than 500K parameters.

We construct training data by first fine-tuning 10 transformer-base models on 3 text corpora, and then extracting features for samples included and not included in fine-tuning. We sample uniformly across the 30 model-dataset combinations during training. The objective for training the attack model is binary cross-entropy, optimized with AdamW~\citep{loshchilov2017decoupled} (learning rate $3 \times 10^{-4}$, batch size 1024) for 30 epochs. We select the checkpoint with the highest AUC on the validation dataset.

\section{Experiments}

We evaluate LT-MIA on held-out model-dataset combinations, with particular focus on transfer to architectures not seen during training.

\subsection{Experimental Setup}

\textit{Data.} Training uses 10 transformer models (listed in Appendix~\ref{app:split}) fine-tuned on 3 text corpora, yielding 30 model-dataset combinations with 18,000 samples each (540,000 total). Evaluation uses 7 held-out transformers (Appendix~\ref{app:split}) plus three non-transformer architectures: Mamba-2.8B (state-space), RWKV-4-3B (linear attention), and RecurrentGemma-2B (gated recurrence). Each is evaluated on 4 datasets: AG News~\citep{zhang2015character}, WikiText-103~\citep{merity2016pointer}, XSum~\citep{narayan2018don}, and Swallow-Code~\citep{fujii2025rewriting}. There is zero overlap between training and evaluation combinations (Appendix~\ref{app:split}).

\textit{Baselines.} We compare against training-free methods that use target-model statistics (Loss~\citep{yeom2018privacy}, Min-K\%++~\citep{zhang2024mink}, Zlib~\citep{carlini2021extracting}) and reference-based methods that compare target to pre-trained outputs: RefLoss~\citep{carlini2022membership} and EZ-MIA~\citep{ilic2026ezmia}, which aggregates probability shifts at positions where the target model errs. All reference-based methods use the pre-trained checkpoint as reference.

\textit{Metrics.} We report AUC and true positive rate at 1\% and 0.1\% false positive rate. Low-FPR metrics are critical because realistic attacks query many candidates, making tail calibration essential.

\subsection{Results}

\begin{table}[t]
    \caption{AUC by method and architecture. All methods are evaluated on held-out combinations, only LT-MIA involves training. Means for transformers computed over 7 models $\times$ 4 datasets, non-transformer means computed over 4 datasets each.}
    \label{tab:main_results}
    \centering
    \begin{tabular}{lccccc}
        \toprule
        Method & Transformer & Mamba & RWKV & RecurrentGemma & Mean \\
        \midrule
        Loss & 0.622 & 0.867 & 0.786 & 0.860 & 0.784 \\
        Min-K\%++ & 0.585 & 0.814 & 0.706 & 0.813 & 0.730 \\
        Zlib & 0.611 & 0.837 & 0.761 & 0.836 & 0.761 \\
        RefLoss & 0.760 & 0.892 & 0.915 & 0.800 & 0.842 \\
        EZ-MIA & 0.887 & 0.931 & 0.971 & 0.915 & 0.926 \\
        \midrule
        \textbf{LT-MIA} & \textbf{0.908} & \textbf{0.963} & \textbf{0.972} & \textbf{0.936} & \textbf{0.945} \\
        \bottomrule
    \end{tabular}
\end{table}

\begin{table}[t]
    \caption{True positive rate at 1\% and 0.1\% false positive rate across 4 datasets.}
    \label{tab:tpr}
    \centering
    \small
    \begin{tabular}{lccccc}
        \toprule
        \multicolumn{6}{c}{\textbf{TPR @ 1\% FPR}} \\
        \midrule
        Method & Transformer & Mamba & RWKV & RecurrentGemma & Mean \\
        \midrule
        Loss & 0.036 & 0.356 & 0.108 & 0.185 & 0.171 \\
        RefLoss & 0.056 & 0.346 & 0.308 & 0.094 & 0.201 \\
        EZ-MIA & 0.230 & 0.526 & 0.439 & 0.273 & 0.367 \\
        \textbf{LT-MIA} & \textbf{0.375} & \textbf{0.569} & \textbf{0.593} & \textbf{0.399} & \textbf{0.484} \\
        \midrule
        \multicolumn{6}{c}{\textbf{TPR @ 0.1\% FPR}} \\
        \midrule
        Method & Transformer & Mamba & RWKV & RecurrentGemma & Mean \\
        \midrule
        Loss & 0.009 & 0.151 & 0.027 & 0.074 & 0.065 \\
        RefLoss & 0.014 & 0.148 & 0.056 & 0.046 & 0.066 \\
        EZ-MIA & 0.061 & \textbf{0.240} & 0.277 & \textbf{0.226} & 0.201 \\
        \textbf{LT-MIA} & \textbf{0.171} & \textbf{0.240} & \textbf{0.292} & 0.166 & \textbf{0.217} \\
        \bottomrule
    \end{tabular}
\end{table}

Tables~\ref{tab:main_results} and~\ref{tab:tpr} present results across four architectural families.

\textit{LT-MIA significantly outperforms all baselines on every family (vs. EZ-MIA: Wilcoxon signed-rank, p < 0.001).} On held-out transformers, LT-MIA achieves 0.908 mean AUC, exceeding EZ-MIA (0.887) by 2.1 points and RefLoss (0.760) by 14.8 points. The same pattern holds for Mamba (0.963 vs.\ 0.931), RWKV (0.972 vs.\ 0.971), and RecurrentGemma (0.936 vs.\ 0.915).

\textit{All likelihood-based methods transfer.} Transfer to non-transformer architectures is not an artifact of LT-MIA's design. Loss achieves 0.867 AUC on Mamba, RefLoss achieves 0.892, while EZ-MIA reaches 0.931 on Mamba and 0.971 on RWKV. The memorization signature exists independently of detection method: LT-MIA captures it most effectively, but the signal is architecture-invariant.

\textit{Non-transformer performance exceeds transformer performance.} For every method in Table~\ref{tab:main_results}, mean AUC on the three non-transformer families exceeds mean AUC on held-out transformers. LT-MIA achieves 0.957 mean AUC on non-transformers versus 0.908 on transformers. This pattern holds for Loss (0.838 vs.\ 0.622), RefLoss (0.869 vs.\ 0.760), and EZ-MIA (0.939 vs.\ 0.887). To control for training regime and scale, we restrict the transformer comparison to fully fine-tuned models in the 2–7B range (Pythia-2.8B, Pythia-6.9B, Llama-2-7B; mean 0.914 AUC); the gap narrows only slightly, from 4.9 to 4.3 points.

\textit{LT-MIA's advantage grows at strict thresholds.} Table~\ref{tab:tpr} shows that the gap between LT-MIA and baselines widens at low false positive rates. On transformers, LT-MIA achieves 1.6$\times$ higher TPR@1\% than EZ-MIA (0.375 vs.\ 0.230) and 2.8$\times$ higher TPR@0.1\% (0.171 vs.\ 0.061). EZ-MIA's TPR@1\% collapses to near-zero on WikiText-103 for several models while LT-MIA remains robust (Appendix~\ref{app:scatter}).

\textit{Domain transfer.} LT-MIA achieves 0.865 mean AUC on Swallow-Code despite training only on natural language text, demonstrating robustness beyond architectural transfer.

\section{Analysis}
\label{sec:analysis}

The experiments demonstrate that LT-MIA transfers across architectures, but they do not explain why. We analyze three factors: feature importance, training diversity, and sequence modeling.

\subsection{Feature Importance is Consistent Across Families}

\begin{figure}[t]
    \centering
    \includegraphics[width=\linewidth]{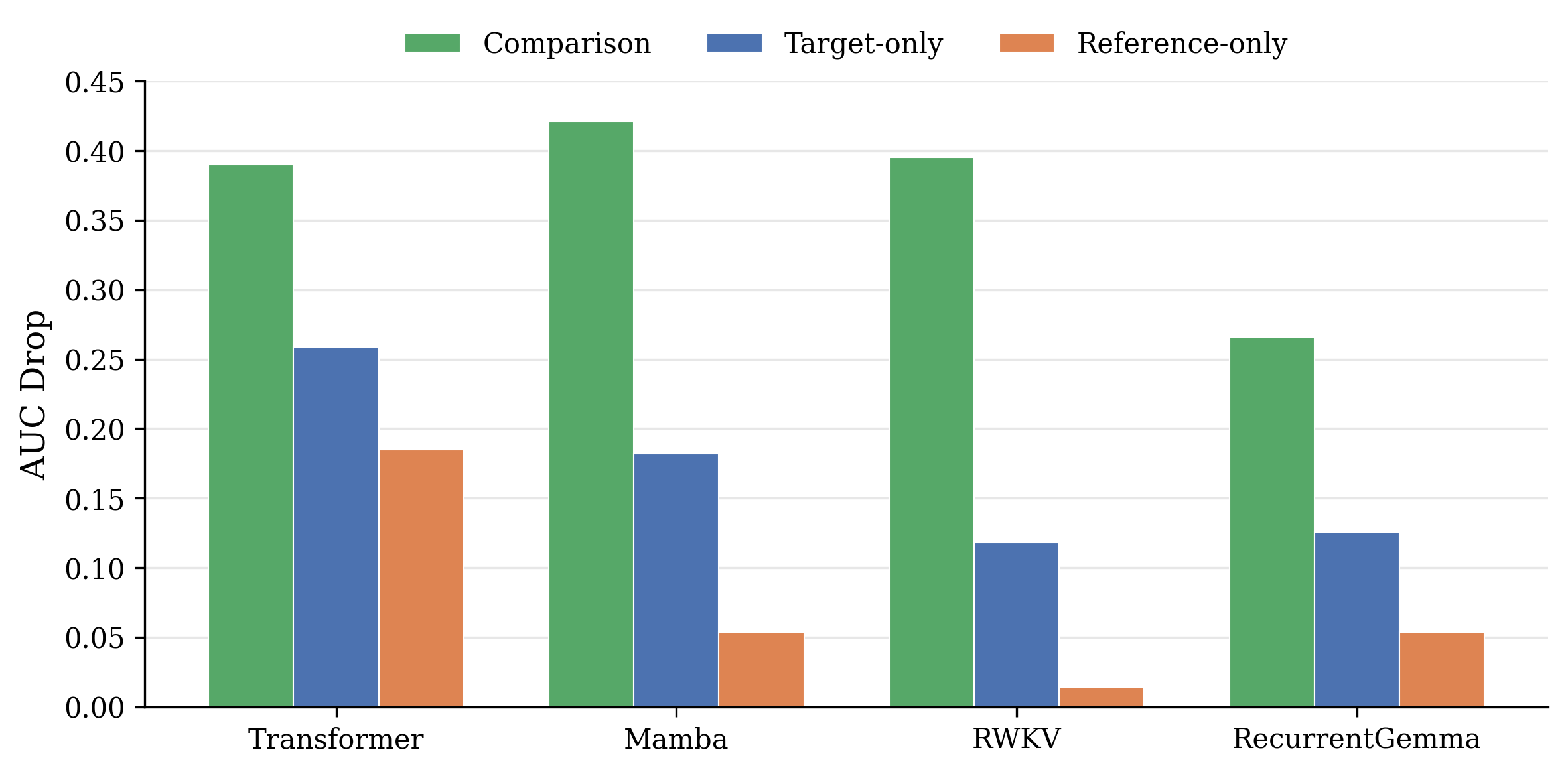}
    \caption{Feature importance measured as AUC drop when each feature group is ablated. Comparison features (relating target to reference model outputs) dominate across all four architectural families. The consistent hierarchy (Comparison $>$ Target-only $>$ Reference-only) confirms the membership signal is relational: what matters is how fine-tuning changed the model, not the model's behavior in isolation.}
    \label{fig:feature_importance}
\end{figure}

Figure~\ref{fig:feature_importance} shows permutation importance, the AUC drop when each feature group's values are randomly shuffled, breaking its association with the label. Comparison features (those relating target to reference model outputs, Section~\ref{sec:features}) dominate for every architectural family, contributing substantially more than target-only or reference-only features. This hierarchy holds for transformers (0.390 vs.\ 0.259 vs.\ 0.185), Mamba (0.421 vs.\ 0.182 vs.\ 0.054), RWKV (0.395 vs.\ 0.118 vs.\ 0.014), and RecurrentGemma (0.266 vs.\ 0.126 vs.\ 0.054).

The consistency is striking: four families with no shared computational mechanisms exhibit the same feature importance pattern. The membership signal is fundamentally relational: what matters is not the target model's behavior in isolation, but how it has changed from its pre-trained state. This relational signal exists in any model trained on cross-entropy loss, explaining why transfer succeeds.

\subsection{Diversity Enables Transfer}

A classifier trained on one model-dataset combination might learn artifacts specific to that setting: tokenizer idiosyncrasies, architectural biases in logit scales, or domain-specific patterns. Training on diverse combinations should filter out such artifacts, leaving only signal that generalizes.

\begin{figure}[t]
    \centering
    \includegraphics[width=0.8\linewidth]{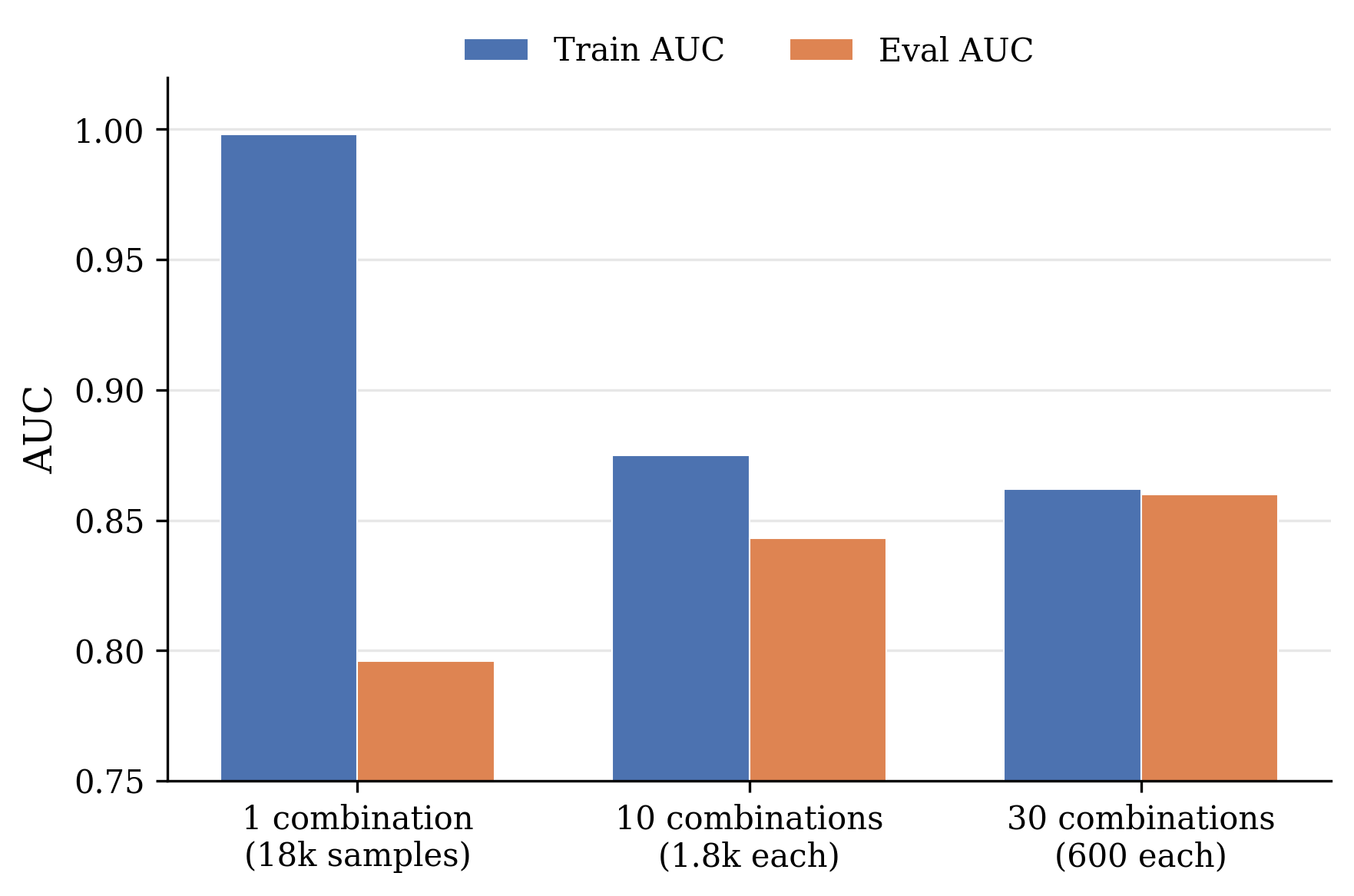}
    \caption{Effect of training diversity on generalization, with total samples fixed at 18,000. ``Train AUC'' is evaluated on held-out samples from training model-dataset combinations (in-distribution), while ``Eval AUC'' is evaluated on entirely different model-dataset combinations (out-of-distribution). All metrics use samples never seen during classifier training. Training on one combination yields near-perfect in-distribution AUC (0.998) but poor transfer (0.796). With 30 combinations, the gap shrinks from 20.2 points to 0.2 points.}
    \label{fig:diversity}
\end{figure}

Figure~\ref{fig:diversity} confirms this hypothesis. Training on a single combination overfits to model-specific artifacts, evident in the 20.2-point generalization gap. With 30 combinations, the classifier is forced to find signal that works across many transformers, filtering out transformer-specific patterns and retaining only the universal memorization signature. This explains why transfer to non-transformers succeeds despite training exclusively on transformers.
\subsection{Sequence Modeling Captures Positional Structure}

Prior methods aggregate per-token statistics into scalar features before classification, discarding positional information. We test whether sequence modeling captures patterns that aggregation misses.

\begin{table}[t]
    \caption{Classifier architecture comparison on held-out combinations. Sequence modeling contributes 5.0 AUC points over pooling.}
    \label{tab:classifier}
    \centering
    \begin{tabular}{lcc}
        \toprule
        Classifier & Train AUC & Eval AUC \\
        \midrule
        Transformer (sequence) & 0.914 & \textbf{0.925} \\
        MLP (flattened) & 0.885 & 0.877 \\
        MLP (mean pooled) & 0.885 & 0.875 \\
        Logistic Regression & 0.838 & 0.839 \\
        \bottomrule
    \end{tabular}
\end{table}

Table~\ref{tab:classifier} compares four classifier architectures trained on identical features. The transformer achieves 5.0 points higher AUC than the mean-pooled MLP (0.925 vs.\ 0.875). Notably, the flattened MLP (0.877) performs comparably to the mean-pooled MLP despite retaining positional information, suggesting the transformer's advantage stems from attention's ability to weight positions adaptively rather than position preservation alone.

\section{Discussion}

\subsection{Moving MIA into the Deep Learning Era}

All prior membership inference methods for language models are hand-crafted heuristics. Loss thresholding~\citep{yeom2018privacy}, Min-K\%~\citep{shi2024detecting}, Zlib normalization~\citep{carlini2021extracting}, RefLoss~\citep{carlini2022membership}, neighborhood comparison~\citep{mattern2023membership}, and EZ-MIA~\citep{ilic2026ezmia} each design a statistic by intuition and threshold it. Performance is bounded by the designer's insight into what signals membership.

Learned attacks exist for image classifiers~\citep{shokri2017membership, salem2019mlleaks}, but the shadow model paradigm requires training many models with known membership, which is prohibitively expensive at LLM scale. The implicit assumption has been that learned MIA requires shadow models, and therefore learned MIA cannot scale to language models.

We remove this bottleneck with a simple observation: MIA has unlimited labeled training data by construction. Fine-tuning any model on any corpus yields perfect membership labels. No shadow models are needed, only feature extraction from existing fine-tuned models is required. The cost is forward passes, not training. This insight enables standard deep learning benefits: learning what matters rather than designing it, generalization through training diversity combined with scale, and transfer to settings never seen during training.

\subsection{Scaling}

Our current setup uses a 500K parameter classifier, 30 model-dataset combinations, and 154 features. All of these can be scaled. The deep learning playbook predicts performance gains from more training combinations, larger classifier capacity, and richer feature representations. Our diversity ablation (Section~\ref{sec:analysis}) already shows that 30 combinations substantially outperform 1 combination at fixed data volume, but the gains from 100 or 1000 combinations remain unexplored. This paper demonstrates proof-of-concept, and the performance ceiling is likely higher; finding the scaling laws that govern learned membership inference remains an open question.

\subsection{Beyond Cross-Entropy Fine-Tuning}

We validated the memorization signature on supervised fine-tuning with cross-entropy loss. Whether it persists under other training paradigms remains open. RLHF~\citep{ouyang2022training}, DPO~\citep{rafailov2023direct}, instruction tuning, and continual pretraining all modify the output distribution through gradient descent, so if the signature arises from optimization pressure toward training data, it may generalize beyond cross-entropy. Pretraining presents a different challenge: \citet{duan2024membership} showed existing methods achieve near-random performance, likely due to weak per-example memorization from large datasets and few epochs. A learned approach might extract signal that heuristics miss, or the signal may genuinely be weaker, and distinguishing these two possibilities is an important direction for future work.

\subsection{Defense Implications}

If the memorization signature is intrinsic to gradient-based training, can it be defended? Differential privacy provides formal guarantees but severely degrades utility at the noise levels required for meaningful protection~\citep{abadi2016deep, carlini2022membership}. An open question is whether training modifications can hide the signature without destroying model performance, or whether this represents a fundamental tradeoff between learning from data and leaking information about that data. Our finding that the signature transfers across architectural families suggests that defenses targeting specific computational mechanisms (attention patterns, state dynamics) are unlikely to succeed, though further research is necessary.

\section{Limitations}

Our method struggles when the reference model already assigns high likelihood to the target text. On WikiText-103 with large models (Llama-2-7B: 0.781 AUC, Pythia-6.9B: 0.796 AUC), performance drops below our 0.908 average. WikiText-103 likely appears in these models' pretraining corpora: when the reference model already ``knows'' a text, fine-tuning produces a smaller likelihood shift, weakening the signal. This is fundamental to reference-based membership inference: these methods detect the change induced by training, and when training induces little change, detection is inherently harder.

Several practical constraints define the method's scope. Like all reference-based attacks, LT-MIA requires black-box access to a pre-trained checkpoint with the same architecture as the target, which is available for open-weight models but not closed APIs. Our experiments validate the method on fine-tuning. Whether it extends to pretraining membership or RLHF remains untested. Transfer to code achieves 0.865 AUC versus 0.927 on text, still above all baselines but with reduced performance.

\section{Broader Impact}
\label{sec:broader_impact}

Membership inference research is inherently dual-use. The same method enabling a copyright holder to verify unauthorized use of their content also enables an adversary to infer sensitive training data membership. We believe this work favors defenders, but acknowledge the tension.

\textit{Beneficial applications.} LT-MIA enables copyright holders to check whether their content was used for fine-tuning, supports organizational compliance with data use agreements, allows model developers to audit memorization before release, and provides regulators a tool for enforcement (e.g., GDPR Article 15 right-of-access requests).

\textit{Why characterization helps defenders.} The cross-architecture vulnerability exists independent of our method: Loss achieves 0.867 AUC on Mamba, EZ-MIA reaches 0.931, all without training. We did not introduce this transfer, just characterized it. Knowing the signature is architecture-invariant tells defenders not to invest in architectural migration (switching to state-space models will not help) and to focus instead on training objective modifications. A single audit tool can now check any model family. Additionally, our method's constraints (reference model access, full vocabulary logits, fine-tuning only) limit applicability to open-weight models, while closed APIs remain unaffected.

\textit{Scientific implications.} Architecture-invariance reveals that memorization likely arises from cross-entropy, not architecture, and understanding this signature is prerequisite to hiding it.

\bibliographystyle{plainnat}
\bibliography{bibliography}

@inproceedings{vaswani2017attention,
  title={Attention is All You Need},
  author={Vaswani, Ashish and Shazeer, Noam and Parmar, Niki and Uszkoreit, Jakob and Jones, Llion and Gomez, Aidan N and Kaiser, {\L}ukasz and Polosukhin, Illia},
  booktitle={Advances in Neural Information Processing Systems},
  volume={30},
  year={2017}
}

@article{gu2023mamba,
  title={Mamba: Linear-Time Sequence Modeling with Selective State Spaces},
  author={Gu, Albert and Dao, Tri},
  journal={arXiv preprint arXiv:2312.00752},
  year={2023}
}

@inproceedings{peng2023rwkv,
  title={{RWKV}: Reinventing {RNN}s for the Transformer Era},
  author={Peng, Bo and Alcaide, Eric and Anthony, Quentin and Albalak, Alon and Arcadinho, Samuel and Biderman, Stella and Cao, Huanqi and Cheng, Xin and Chung, Michael and Derczynski, Leon and Du, Xingjian and Grella, Matteo and Gv, Kranthi and He, Xuzheng and Hou, Haowen and Kazienko, Przemyslaw and Kocon, Jan and Kong, Jiaming and Koptyra, Bart{\l}omiej and Lau, Hayden and Lin, Jiaju and Mantri, Krishna Sri Ipsit and Mom, Ferdinand and Saito, Atsushi and Song, Guangyu and Tang, Xiangru and Wind, Johan and Wo{\'z}niak, Stanis{\l}aw and Zhang, Zhenyuan and Zhou, Qinghua and Zhu, Jian and Zhu, Rui-Jie},
  booktitle={Findings of the Association for Computational Linguistics: EMNLP 2023},
  pages={14048--14077},
  year={2023}
}

@article{botev2024recurrentgemma,
  title={RecurrentGemma: Moving Past Transformers for Efficient Open Language Models},
  author={Botev, Aleksandar and De, Soham and Smith, Samuel L and Fernando, Anushan and Muraru, George-Cristian and Haroun, Ruba and Berrada, Leonard and Pascanu, Razvan and Sessa, Pier Giuseppe and Dadashi, Robert and Hussenot, L{\'e}onard and Ferret, Johan and Girgin, Sertan and Bachem, Olivier and Andreev, Alek and Kenealy, Kathleen and Mesnard, Thomas and Hardin, Cassidy and Bhupatiraju, Surya and Pathak, Shreya and Sifre, Laurent and Rivi{\`e}re, Morgane and Kale, Mihir Sanjay and Love, Juliette and Tafti, Pouya and Joulin, Armand and Fiedel, Noah and Senter, Evan and Chen, Yutian and Srinivasan, Srivatsan and Desjardins, Guillaume and Budden, David and Doucet, Arnaud and Vikram, Sharad and Paszke, Adam and Gale, Trevor and Borgeaud, Sebastian and Chen, Charlie and Brock, Andy and Paterson, Antonia and Brennan, Jenny and Risdal, Meg and Gundluru, Raj and Devanathan, Nesh and Mooney, Paul and Chauhan, Nilay and Culliton, Phil and Martins, Luiz Gustavo and Bandy, Elisa and Huntsperger, David and Cameron, Glenn and Zucker, Arthur and Warkentin, Tris and Peran, Ludovic and Giang, Minh and Ghahramani, Zoubin and Farabet, Cl{\'e}ment and Kavukcuoglu, Koray and Hassabis, Demis and Hadsell, Raia and Teh, Yee Whye and de Freitas, Nando},
  journal={arXiv preprint arXiv:2404.07839},
  year={2024}
}

@article{radford2019language,
  title={Language Models are Unsupervised Multitask Learners},
  author={Radford, Alec and Wu, Jeffrey and Child, Rewon and Luan, David and Amodei, Dario and Sutskever, Ilya},
  journal={OpenAI blog},
  volume={1},
  number={8},
  pages={9},
  year={2019}
}

@inproceedings{biderman2023pythia,
  title={Pythia: A Suite for Analyzing Large Language Models Across Training and Scaling},
  author={Biderman, Stella and Schoelkopf, Hailey and Anthony, Quentin and Bradley, Herbie and O'Brien, Kyle and Hallahan, Eric and Khan, Mohammad Aflah and Purohit, Shivanshu and Prashanth, USVSN Sai and Raff, Edward and Skowron, Aviya and Sutawika, Lintang and van der Wal, Oskar},
  booktitle={International Conference on Machine Learning},
  pages={2397--2430},
  year={2023},
  organization={PMLR}
}

@article{sanh2019distilbert,
  title={DistilBERT, a Distilled Version of BERT: Smaller, Faster, Cheaper and Lighter},
  author={Sanh, Victor and Debut, Lysandre and Chaumond, Julien and Wolf, Thomas},
  journal={arXiv preprint arXiv:1910.01108},
  year={2019}
}

@misc{gptj,
  title={{GPT-J-6B}: A 6 Billion Parameter Autoregressive Language Model},
  author={Wang, Ben and Komatsuzaki, Aran},
  howpublished={\url{https://github.com/kingoflolz/mesh-transformer-jax}},
  year={2021}
}

@article{team2024gemma,
  title={Gemma: Open Models Based on Gemini Research and Technology},
  author={{Gemma Team}},
  journal={arXiv preprint arXiv:2403.08295},
  year={2024}
}

@article{bai2023qwen,
  title={Qwen Technical Report},
  author={Bai, Jinze and Bai, Shuai and Chu, Yunfei and Cui, Zeyu and Dang, Kai and Deng, Xiaodong and Fan, Yang and Ge, Wenbin and Han, Yu and Huang, Fei and others},
  journal={arXiv preprint arXiv:2309.16609},
  year={2023}
}

@misc{mosaicml2023mpt,
  title={{MPT-7B}: A New Standard for Open-Source, Commercially Usable LLMs},
  author={{MosaicML NLP Team}},
  howpublished={\url{https://www.mosaicml.com/blog/mpt-7b}},
  year={2023}
}

@article{penedo2023refinedweb,
  title={The {RefinedWeb} Dataset for {Falcon LLM}: Outperforming Curated Corpora with Web Data, and Web Data Only},
  author={Penedo, Guilherme and Malartic, Quentin and Hesslow, Daniel and Cojocaru, Ruxandra and Cappelli, Alessandro and Alobeidli, Hamza and Pannier, Baptiste and Almazrouei, Ebtesam and Launay, Julien},
  journal={arXiv preprint arXiv:2306.01116},
  year={2023}
}

@article{dey2023cerebras,
  title={Cerebras-{GPT}: Open Compute-Optimal Language Models Trained on the Cerebras Wafer-Scale Cluster},
  author={Dey, Nolan and Gosal, Gurpreet and Zhiming and Chen and Khachane, Hemant and Marshall, William and Pathria, Ribhu and Tom, Marvin and Hestness, Joel},
  journal={arXiv preprint arXiv:2304.03208},
  year={2023}
}

@inproceedings{zhang2015character,
  title={Character-level Convolutional Networks for Text Classification},
  author={Zhang, Xiang and Zhao, Junbo and LeCun, Yann},
  booktitle={Advances in Neural Information Processing Systems},
  volume={28},
  year={2015}
}

@article{merity2016pointer,
  title={Pointer Sentinel Mixture Models},
  author={Merity, Stephen and Xiong, Caiming and Bradbury, James and Socher, Richard},
  journal={arXiv preprint arXiv:1609.07843},
  year={2016}
}

@inproceedings{narayan2018don,
  title={Don't Give Me the Details, Just the Summary! Topic-Aware Convolutional Neural Networks for Extreme Summarization},
  author={Narayan, Shashi and Cohen, Shay B and Lapata, Mirella},
  booktitle={Proceedings of the 2018 Conference on Empirical Methods in Natural Language Processing},
  pages={1797--1807},
  year={2018}
}

@inproceedings{hermann2015teaching,
  title={Teaching Machines to Read and Comprehend},
  author={Hermann, Karl Moritz and Kocisky, Tomas and Grefenstette, Edward and Espeholt, Lasse and Kay, Will and Suleyman, Mustafa and Blunsom, Phil},
  booktitle={Advances in Neural Information Processing Systems},
  volume={28},
  year={2015}
}

@article{misra2022news,
  title={News Category Dataset},
  author={Misra, Rishabh},
  journal={arXiv preprint arXiv:2209.11429},
  year={2022}
}

@misc{wikidump,
  author = {{Wikimedia Foundation}},
  title = {Wikimedia Downloads},
  year = {2024},
  howpublished = {\url{https://dumps.wikimedia.org}}
}

@misc{fujii2025rewriting,
  title={Rewriting Pre-Training Data Boosts LLM Performance in Math and Code},
  author={Kazuki Fujii and Yukito Tajima and Sakae Mizuki and Hinari Shimada and Taihei Shiotani and Koshiro Saito and Masanari Ohi and Masaki Kawamura and Taishi Nakamura and Takumi Okamoto and Shigeki Ishida and Kakeru Hattori and Youmi Ma and Hiroya Takamura and Rio Yokota and Naoaki Okazaki},
  year={2025},
  eprint={2505.02881},
  archivePrefix={arXiv},
  primaryClass={cs.LG}
}

@article{ouyang2022training,
  title={Training Language Models to Follow Instructions with Human Feedback},
  author={Ouyang, Long and Wu, Jeffrey and Jiang, Xu and Almeida, Diogo and Wainwright, Carroll and Mishkin, Pamela and Zhang, Chong and Agarwal, Sandhini and Slama, Katarina and Ray, Alex and others},
  journal={Advances in Neural Information Processing Systems},
  volume={35},
  pages={27730--27744},
  year={2022}
}

@article{rafailov2023direct,
  title={Direct Preference Optimization: Your Language Model is Secretly a Reward Model},
  author={Rafailov, Rafael and Sharma, Archit and Mitchell, Eric and Ermon, Stefano and Manning, Christopher D and Finn, Chelsea},
  journal={arXiv preprint arXiv:2305.18290},
  year={2023}
}

@inproceedings{abadi2016deep,
  title={Deep Learning with Differential Privacy},
  author={Abadi, Martin and Chu, Andy and Goodfellow, Ian and McMahan, H Brendan and Mironov, Ilya and Talwar, Kunal and Zhang, Li},
  booktitle={Proceedings of the 2016 ACM SIGSAC Conference on Computer and Communications Security},
  pages={308--318},
  year={2016}
}

@inproceedings{shokri2017membership,
  title={Membership Inference Attacks Against Machine Learning Models},
  author={Shokri, Reza and Stronati, Marco and Song, Congzheng and Shmatikov, Vitaly},
  booktitle={IEEE Symposium on Security and Privacy},
  pages={3--18},
  year={2017}
}

@inproceedings{yeom2018privacy,
  title={Privacy Risk in Machine Learning: Analyzing the Connection to Overfitting},
  author={Yeom, Samuel and Giacomelli, Irene and Fredrikson, Matt and Jha, Somesh},
  booktitle={IEEE Computer Security Foundations Symposium},
  pages={268--282},
  year={2018}
}

@inproceedings{feldman2020does,
  title={Does Learning Require Memorization? A Short Tale about a Long Tail},
  author={Feldman, Vitaly},
  booktitle={Symposium on Theory of Computing},
  pages={954--959},
  year={2020}
}

@inproceedings{carlini2021extracting,
  title={Extracting Training Data from Large Language Models},
  author={Carlini, Nicholas and Tramer, Florian and Wallace, Eric and Jagielski, Matthew and Herbert-Voss, Ariel and Lee, Katherine and Roberts, Adam and Brown, Tom and Song, Dawn and Erlingsson, Ulfar and others},
  booktitle={USENIX Security Symposium},
  pages={2633--2650},
  year={2021}
}

@inproceedings{carlini2022membership,
  title={Membership Inference Attacks From First Principles},
  author={Carlini, Nicholas and Chien, Steve and Nasr, Milad and Song, Shuang and Terzis, Andreas and Tramer, Florian},
  booktitle={IEEE Symposium on Security and Privacy},
  pages={1897--1914},
  year={2022}
}

@article{shi2024detecting,
  title={Detecting Pretraining Data from Large Language Models},
  author={Shi, Weijia and Ajith, Anirudh and Xia, Mengzhou and Huang, Yangsibo and Liu, Daogao and Blevins, Terra and Chen, Danqi and Zettlemoyer, Luke},
  journal={arXiv preprint arXiv:2310.16789},
  year={2024}
}

@article{zhang2024mink,
  title={Min-K\%++: Improved Baseline for Detecting Pre-Training Data from Large Language Models},
  author={Zhang, Jingyang and Sun, Jingwei and Yeats, Eric and Ouyang, Yang and Kuo, Martin and Zhang, Jianyi and Yang, Hao and Li, Hai},
  journal={arXiv preprint arXiv:2404.02936},
  year={2024}
}

@article{ilic2026ezmia,
  title={Powerful Training-Free Membership Inference Against Autoregressive Language Models},
  author={Ili{\'c}, David and Stanojevi{\'c}, David and Cvejoski, Kostadin},
  journal={arXiv preprint arXiv:2601.12104},
  year={2026}
}

@article{duan2024membership,
  title={Do Membership Inference Attacks Work on Large Language Models?},
  author={Duan, Michael and Suri, Anshuman and Mireshghallah, Niloofar and Min, Sewon and Shi, Weijia and Zettlemoyer, Luke and Tsvetkov, Yulia and Choi, Yejin and Evans, David and Hajishirzi, Hannaneh},
  journal={arXiv preprint arXiv:2402.07841},
  year={2024}
}

@inproceedings{salem2019mlleaks,
  title={{ML-Leaks}: Model and Data Independent Membership Inference Attacks and Defenses on Machine Learning Models},
  author={Salem, Ahmed and Zhang, Yang and Humbert, Mathias and Berrang, Pascal and Fritz, Mario and Backes, Michael},
  booktitle={Network and Distributed Systems Security Symposium},
  year={2019}
}

@inproceedings{mattern2023membership,
  title={Membership Inference Attacks against Language Models via Neighbourhood Comparison},
  author={Mattern, Justus and Mireshghallah, Fatemehsadat and Jin, Zhijing and Sch{\"o}lkopf, Bernhard and Sachan, Mrinmaya and Berg-Kirkpatrick, Taylor},
  booktitle={Findings of the Association for Computational Linguistics: ACL 2023},
  pages={11330--11343},
  year={2023}
}

@inproceedings{loshchilov2017decoupled,
  title={Decoupled Weight Decay Regularization},
  author={Loshchilov, Ilya and Hutter, Frank},
  booktitle={International Conference on Learning Representations},
  year={2019}
}

@article{chen2026wbc,
  title={Window-based Membership Inference Attacks Against Fine-tuned Large Language Models},
  author={Chen, Yuetian and Du, Yuntao and Zhang, Kaiyuan and Kundu, Ashish and Fleming, Charles and Ribeiro, Bruno and Li, Ninghui},
  journal={arXiv preprint arXiv:2601.02751},
  year={2026}
}

\clearpage
\appendix

\section{Full Results}
\label{app:full_results}

Tables~\ref{tab:full_transformer_auc}--\ref{tab:full_nontransformer_tpr01} present complete results for all model-dataset combinations.

\begin{table}[h]
\caption{Full results on held-out transformers: AUC.}
\label{tab:full_transformer_auc}
\centering
\small
\begin{tabular}{llcccccc}
\toprule
Model & Dataset & Loss & Min-K\%++ & Zlib & RefLoss & EZ-MIA & LT-MIA \\
\midrule
GPT-2 & AG News & 0.745 & 0.704 & 0.717 & 0.790 & \textbf{0.960} & 0.945 \\
GPT-2 & WikiText & 0.745 & 0.696 & 0.713 & 0.814 & 0.971 & \textbf{0.980} \\
GPT-2 & XSum & 0.768 & 0.719 & 0.760 & 0.956 & \textbf{0.994} & 0.991 \\
GPT-2 & Code & 0.618 & 0.608 & 0.618 & 0.608 & \textbf{0.795} & 0.761 \\
Pythia-2.8B & AG News & 0.603 & 0.565 & 0.584 & 0.740 & 0.872 & \textbf{0.911} \\
Pythia-2.8B & WikiText & 0.585 & 0.558 & 0.583 & 0.645 & 0.867 & \textbf{0.944} \\
Pythia-2.8B & XSum & 0.618 & 0.581 & 0.609 & 0.862 & 0.911 & \textbf{0.975} \\
Pythia-2.8B & Code & 0.596 & 0.552 & 0.586 & 0.717 & 0.857 & \textbf{0.865} \\
Pythia-6.9B & AG News & 0.638 & 0.597 & 0.615 & 0.796 & 0.917 & \textbf{0.942} \\
Pythia-6.9B & WikiText & 0.617 & 0.582 & 0.606 & 0.684 & \textbf{0.911} & 0.796 \\
Pythia-6.9B & XSum & 0.649 & 0.607 & 0.638 & 0.902 & 0.957 & \textbf{0.990} \\
Pythia-6.9B & Code & 0.626 & 0.573 & 0.612 & 0.761 & 0.909 & \textbf{0.911} \\
OPT-2.7B & AG News & 0.581 & 0.554 & 0.564 & 0.740 & 0.879 & \textbf{0.908} \\
OPT-2.7B & WikiText & 0.578 & 0.556 & 0.578 & 0.623 & 0.852 & \textbf{0.862} \\
OPT-2.7B & XSum & 0.628 & 0.592 & 0.621 & 0.899 & 0.940 & \textbf{0.979} \\
OPT-2.7B & Code & 0.561 & 0.540 & 0.562 & 0.605 & 0.692 & \textbf{0.751} \\
Llama-2-7B & AG News & 0.627 & 0.575 & 0.599 & 0.815 & 0.917 & \textbf{0.948} \\
Llama-2-7B & WikiText & 0.529 & 0.520 & 0.540 & 0.568 & 0.733 & \textbf{0.781} \\
Llama-2-7B & XSum & 0.671 & 0.620 & 0.671 & 0.930 & 0.972 & \textbf{0.988} \\
Llama-2-7B & Code & 0.624 & 0.571 & 0.609 & 0.784 & 0.902 & \textbf{0.912} \\
Phi-2 & AG News & 0.585 & 0.556 & 0.568 & 0.698 & 0.845 & \textbf{0.886} \\
Phi-2 & WikiText & 0.573 & 0.548 & 0.572 & 0.620 & \textbf{0.859} & 0.823 \\
Phi-2 & XSum & 0.624 & 0.592 & 0.614 & 0.792 & 0.889 & \textbf{0.952} \\
Phi-2 & Code & 0.573 & 0.534 & 0.562 & 0.732 & 0.824 & \textbf{0.851} \\
Mistral-7B & AG News & 0.595 & 0.551 & 0.573 & 0.787 & 0.883 & \textbf{0.916} \\
Mistral-7B & WikiText & 0.599 & 0.569 & 0.601 & 0.689 & 0.910 & \textbf{0.966} \\
Mistral-7B & XSum & 0.660 & 0.615 & 0.660 & 0.933 & 0.952 & \textbf{0.989} \\
Mistral-7B & Code & 0.594 & 0.547 & 0.578 & 0.777 & 0.856 & \textbf{0.903} \\
\bottomrule
\end{tabular}
\end{table}

\begin{table}[h]
\caption{Full results on held-out transformers: TPR @ 1\% FPR.}
\label{tab:full_transformer_tpr1}
\centering
\small
\begin{tabular}{llcccccc}
\toprule
Model & Dataset & Loss & Min-K\%++ & Zlib & RefLoss & EZ-MIA & LT-MIA \\
\midrule
GPT-2 & AG News & 0.018 & 0.026 & 0.014 & 0.020 & \textbf{0.408} & 0.274 \\
GPT-2 & WikiText & 0.084 & 0.024 & 0.092 & 0.018 & 0.058 & \textbf{0.632} \\
GPT-2 & XSum & 0.122 & 0.100 & 0.104 & 0.090 & \textbf{0.868} & 0.756 \\
GPT-2 & Code & 0.024 & 0.010 & 0.010 & 0.010 & 0.030 & \textbf{0.054} \\
Pythia-2.8B & AG News & 0.014 & 0.020 & 0.012 & 0.042 & 0.132 & \textbf{0.257} \\
Pythia-2.8B & WikiText & 0.014 & 0.012 & 0.020 & 0.010 & 0.016 & \textbf{0.492} \\
Pythia-2.8B & XSum & 0.036 & 0.022 & 0.030 & 0.120 & 0.430 & \textbf{0.653} \\
Pythia-2.8B & Code & 0.034 & 0.040 & 0.036 & 0.028 & \textbf{0.180} & 0.129 \\
Pythia-6.9B & AG News & 0.012 & 0.024 & 0.012 & 0.052 & 0.208 & \textbf{0.319} \\
Pythia-6.9B & WikiText & 0.032 & 0.016 & 0.022 & 0.010 & 0.036 & \textbf{0.137} \\
Pythia-6.9B & XSum & 0.058 & 0.036 & 0.056 & 0.168 & 0.670 & \textbf{0.826} \\
Pythia-6.9B & Code & 0.038 & 0.042 & 0.040 & 0.078 & \textbf{0.250} & 0.217 \\
OPT-2.7B & AG News & 0.020 & 0.018 & 0.010 & 0.036 & 0.218 & \textbf{0.264} \\
OPT-2.7B & WikiText & 0.010 & 0.012 & 0.024 & 0.010 & 0.000 & \textbf{0.262} \\
OPT-2.7B & XSum & 0.040 & 0.040 & 0.022 & 0.050 & 0.350 & \textbf{0.696} \\
OPT-2.7B & Code & 0.018 & 0.028 & 0.010 & 0.014 & 0.024 & \textbf{0.064} \\
Llama-2-7B & AG News & 0.024 & 0.016 & 0.012 & 0.072 & 0.252 & \textbf{0.317} \\
Llama-2-7B & WikiText & 0.020 & 0.018 & 0.022 & 0.012 & 0.020 & \textbf{0.130} \\
Llama-2-7B & XSum & 0.044 & 0.040 & 0.058 & 0.136 & 0.628 & \textbf{0.809} \\
Llama-2-7B & Code & 0.060 & 0.026 & 0.048 & 0.074 & 0.144 & \textbf{0.228} \\
Phi-2 & AG News & 0.012 & 0.014 & 0.010 & 0.050 & 0.166 & \textbf{0.196} \\
Phi-2 & WikiText & 0.030 & 0.016 & 0.030 & 0.012 & 0.010 & \textbf{0.187} \\
Phi-2 & XSum & 0.024 & 0.016 & 0.010 & 0.056 & 0.252 & \textbf{0.465} \\
Phi-2 & Code & 0.038 & 0.006 & 0.026 & 0.034 & \textbf{0.124} & 0.098 \\
Mistral-7B & AG News & 0.012 & 0.018 & 0.010 & 0.052 & 0.140 & \textbf{0.257} \\
Mistral-7B & WikiText & 0.036 & 0.028 & 0.016 & 0.016 & 0.082 & \textbf{0.666} \\
Mistral-7B & XSum & 0.056 & 0.028 & 0.050 & 0.082 & 0.540 & \textbf{0.848} \\
Mistral-7B & Code & 0.052 & 0.028 & 0.020 & 0.122 & 0.192 & \textbf{0.259} \\
\bottomrule
\end{tabular}
\end{table}

\begin{table}[h]
\caption{Full results on held-out transformers: TPR @ 0.1\% FPR.}
\label{tab:full_transformer_tpr01}
\centering
\small
\begin{tabular}{llcccccc}
\toprule
Model & Dataset & Loss & Min-K\%++ & Zlib & RefLoss & EZ-MIA & LT-MIA \\
\midrule
GPT-2 & AG News & 0.008 & 0.008 & 0.006 & 0.008 & \textbf{0.124} & 0.044 \\
GPT-2 & WikiText & 0.008 & 0.002 & 0.006 & 0.000 & 0.002 & \textbf{0.273} \\
GPT-2 & XSum & 0.032 & 0.048 & 0.016 & 0.016 & \textbf{0.426} & 0.340 \\
GPT-2 & Code & 0.000 & 0.002 & 0.000 & 0.000 & 0.000 & \textbf{0.006} \\
Pythia-2.8B & AG News & 0.006 & 0.008 & 0.006 & 0.000 & 0.034 & \textbf{0.064} \\
Pythia-2.8B & WikiText & 0.002 & 0.000 & 0.000 & 0.004 & 0.000 & \textbf{0.291} \\
Pythia-2.8B & XSum & 0.008 & 0.004 & 0.006 & 0.022 & 0.092 & \textbf{0.262} \\
Pythia-2.8B & Code & 0.018 & 0.018 & 0.000 & 0.002 & 0.036 & \textbf{0.045} \\
Pythia-6.9B & AG News & 0.008 & 0.008 & 0.006 & 0.002 & 0.064 & \textbf{0.073} \\
Pythia-6.9B & WikiText & 0.004 & 0.002 & 0.002 & 0.004 & 0.000 & \textbf{0.035} \\
Pythia-6.9B & XSum & 0.010 & 0.004 & 0.006 & 0.028 & 0.012 & \textbf{0.567} \\
Pythia-6.9B & Code & 0.020 & 0.012 & 0.004 & 0.002 & 0.026 & \textbf{0.051} \\
OPT-2.7B & AG News & 0.008 & 0.008 & 0.006 & 0.020 & 0.012 & \textbf{0.056} \\
OPT-2.7B & WikiText & 0.000 & 0.000 & 0.000 & 0.000 & 0.000 & \textbf{0.101} \\
OPT-2.7B & XSum & 0.006 & 0.002 & 0.006 & 0.010 & 0.066 & \textbf{0.298} \\
OPT-2.7B & Code & 0.002 & 0.004 & 0.002 & 0.002 & 0.000 & \textbf{0.015} \\
Llama-2-7B & AG News & 0.010 & 0.012 & 0.006 & 0.030 & \textbf{0.138} & 0.061 \\
Llama-2-7B & WikiText & 0.004 & 0.006 & 0.006 & 0.002 & 0.000 & \textbf{0.033} \\
Llama-2-7B & XSum & 0.010 & 0.004 & 0.010 & 0.060 & 0.374 & \textbf{0.580} \\
Llama-2-7B & Code & 0.020 & 0.016 & 0.000 & 0.002 & 0.000 & \textbf{0.051} \\
Phi-2 & AG News & 0.008 & 0.012 & 0.006 & 0.010 & 0.000 & \textbf{0.033} \\
Phi-2 & WikiText & 0.004 & 0.004 & 0.004 & 0.008 & 0.000 & \textbf{0.057} \\
Phi-2 & XSum & 0.002 & 0.002 & 0.000 & 0.016 & 0.046 & \textbf{0.190} \\
Phi-2 & Code & 0.012 & 0.002 & 0.010 & 0.004 & \textbf{0.018} & \textbf{0.019} \\
Mistral-7B & AG News & 0.012 & 0.010 & 0.006 & 0.002 & 0.000 & \textbf{0.036} \\
Mistral-7B & WikiText & 0.012 & 0.008 & 0.008 & 0.000 & 0.000 & \textbf{0.428} \\
Mistral-7B & XSum & 0.006 & 0.002 & 0.010 & 0.024 & 0.214 & \textbf{0.667} \\
Mistral-7B & Code & 0.016 & 0.002 & 0.010 & 0.006 & 0.008 & \textbf{0.092} \\
\bottomrule
\end{tabular}
\end{table}

\begin{table}[h]
\caption{Full results on non-transformer architectures: AUC.}
\label{tab:full_nontransformer_auc}
\centering
\small
\begin{tabular}{llcccccc}
\toprule
Model & Dataset & Loss & Min-K\%++ & Zlib & RefLoss & EZ-MIA & LT-MIA \\
\midrule
Mamba-2.8B & AG News & 0.949 & 0.888 & 0.918 & 0.975 & \textbf{0.983} & 0.969 \\
Mamba-2.8B & WikiText & 0.966 & 0.886 & 0.897 & 0.947 & 0.979 & \textbf{0.995} \\
Mamba-2.8B & XSum & 0.988 & 0.949 & 0.977 & 0.999 & 0.999 & \textbf{0.999} \\
Mamba-2.8B & Code & 0.565 & 0.534 & 0.555 & 0.646 & 0.763 & \textbf{0.887} \\
RWKV-3B & AG News & 0.783 & 0.703 & 0.748 & 0.931 & 0.978 & \textbf{0.976} \\
RWKV-3B & WikiText & 0.792 & 0.704 & 0.753 & 0.865 & 0.965 & \textbf{0.986} \\
RWKV-3B & XSum & 0.860 & 0.766 & 0.853 & 0.993 & 0.998 & \textbf{0.998} \\
RWKV-3B & Code & 0.708 & 0.651 & 0.690 & 0.871 & \textbf{0.943} & 0.926 \\
RecurrentGemma-2B & AG News & 0.867 & 0.848 & 0.829 & 0.760 & 0.849 & \textbf{0.924} \\
RecurrentGemma-2B & WikiText & 0.860 & 0.835 & 0.822 & 0.759 & 0.914 & \textbf{0.946} \\
RecurrentGemma-2B & XSum & 0.938 & 0.881 & 0.939 & 0.898 & 0.985 & \textbf{0.988} \\
RecurrentGemma-2B & Code & 0.775 & 0.687 & 0.754 & 0.783 & \textbf{0.911} & 0.887 \\
\bottomrule
\end{tabular}
\end{table}

\begin{table}[h]
\caption{Full results on non-transformer architectures: TPR @ 1\% FPR.}
\label{tab:full_nontransformer_tpr1}
\centering
\small
\begin{tabular}{llcccccc}
\toprule
Model & Dataset & Loss & Min-K\%++ & Zlib & RefLoss & EZ-MIA & LT-MIA \\
\midrule
Mamba-2.8B & AG News & 0.150 & 0.150 & 0.114 & 0.316 & \textbf{0.718} & 0.206 \\
Mamba-2.8B & WikiText & 0.462 & 0.154 & 0.544 & 0.040 & 0.330 & \textbf{0.900} \\
Mamba-2.8B & XSum & 0.774 & 0.518 & 0.800 & 0.998 & 0.986 & \textbf{0.996} \\
Mamba-2.8B & Code & 0.036 & 0.028 & 0.022 & 0.028 & 0.070 & \textbf{0.174} \\
RWKV-3B & AG News & 0.092 & 0.060 & 0.026 & 0.084 & 0.402 & \textbf{0.430} \\
RWKV-3B & WikiText & 0.066 & 0.048 & 0.090 & 0.022 & 0.050 & \textbf{0.624} \\
RWKV-3B & XSum & 0.198 & 0.072 & 0.212 & 0.924 & 0.946 & \textbf{0.981} \\
RWKV-3B & Code & 0.076 & 0.046 & 0.058 & 0.202 & \textbf{0.358} & 0.338 \\
RecurrentGemma-2B & AG News & 0.110 & 0.082 & 0.014 & 0.104 & 0.038 & \textbf{0.392} \\
RecurrentGemma-2B & WikiText & 0.120 & 0.070 & 0.260 & 0.038 & 0.014 & \textbf{0.316} \\
RecurrentGemma-2B & XSum & 0.360 & 0.160 & 0.410 & 0.154 & 0.748 & \textbf{0.750} \\
RecurrentGemma-2B & Code & 0.150 & 0.054 & 0.144 & 0.080 & \textbf{0.290} & 0.138 \\
\bottomrule
\end{tabular}
\end{table}

\begin{table}[h]
\caption{Full results on non-transformer architectures: TPR @ 0.1\% FPR.}
\label{tab:full_nontransformer_tpr01}
\centering
\small
\begin{tabular}{llcccccc}
\toprule
Model & Dataset & Loss & Min-K\%++ & Zlib & RefLoss & EZ-MIA & LT-MIA \\
\midrule
Mamba-2.8B & AG News & 0.044 & \textbf{0.052} & 0.012 & 0.000 & 0.018 & 0.018 \\
Mamba-2.8B & WikiText & 0.070 & 0.020 & 0.038 & 0.002 & 0.004 & \textbf{0.300} \\
Mamba-2.8B & XSum & 0.480 & 0.242 & 0.460 & 0.592 & \textbf{0.888} & 0.582 \\
Mamba-2.8B & Code & 0.010 & 0.018 & 0.002 & 0.002 & 0.048 & \textbf{0.058} \\
RWKV-3B & AG News & 0.010 & 0.010 & 0.008 & 0.002 & \textbf{0.098} & 0.060 \\
RWKV-3B & WikiText & 0.024 & 0.012 & 0.010 & 0.000 & 0.000 & \textbf{0.231} \\
RWKV-3B & XSum & 0.060 & 0.024 & 0.048 & 0.222 & \textbf{0.820} & 0.727 \\
RWKV-3B & Code & 0.014 & 0.020 & 0.014 & 0.002 & \textbf{0.190} & 0.148 \\
RecurrentGemma-2B & AG News & 0.025 & 0.025 & 0.005 & 0.001 & 0.022 & \textbf{0.040} \\
RecurrentGemma-2B & WikiText & 0.033 & 0.013 & 0.022 & 0.001 & 0.001 & \textbf{0.110} \\
RecurrentGemma-2B & XSum & 0.212 & 0.090 & 0.208 & 0.179 & \textbf{0.726} & 0.500 \\
RecurrentGemma-2B & Code & 0.026 & 0.025 & 0.020 & 0.002 & \textbf{0.153} & 0.012 \\
\bottomrule
\end{tabular}
\end{table}

\section{Experimental Configuration}
\label{app:split}

\textbf{Fine-tuning Configuration.} All target models were fine-tuned for 3 epochs with batch size 16, learning rate $5 \times 10^{-5}$, and sequence length 128 tokens. For most transformer models (Mistral-7B, Phi-2, OPT-2.7B, and all training-set transformers except DistilGPT-2), we used LoRA with rank $r=16$, $\alpha=32$, and dropout 0.05, applied to all attention projections and feed-forward layers. Other transformers are fully fine-tuned. Mamba, RWKV, and RecurrentGemma were fully fine-tuned without LoRA. Each model-dataset combination used 10,000 member and 10,000 non-member samples, with 5\% held out for validation and 5\% for testing. The checkpoint with lowest validation loss was selected for feature extraction.

\textbf{Compute.} All experiments were conducted on a single NVIDIA H200 GPU.
We estimate total compute at approximately $9 \times 10^{18}$ FLOPs
(70 fine-tuning runs, feature extraction over 70 model-dataset combinations,
and classifier training), corresponding to approximately 30 GPU-hours
including overhead.

Table~\ref{tab:split} documents the complete separation between training and evaluation. There is zero overlap in models or datasets.

\begin{table}[h]
\caption{Training and evaluation split. Training uses 10 transformer models $\times$ 3 datasets = 30 combinations. Evaluation uses 7 held-out transformers + 3 non-transformers $\times$ 4 datasets = 40 combinations.}
\label{tab:split}
\centering
\footnotesize
\begin{tabular}{@{}lp{11.5cm}@{}}
\toprule
\textbf{Training Models (10)} & DistilGPT-2~\citep{sanh2019distilbert}, GPT-2-XL~\citep{radford2019language}, Pythia-1.4B~\citep{biderman2023pythia}, Cerebras-GPT-2.7B~\citep{dey2023cerebras}, GPT-J-6B~\citep{gptj}, Gemma-2B~\citep{team2024gemma}, Qwen2-1.5B~\citep{bai2023qwen}, MPT-7B~\citep{mosaicml2023mpt}, Falcon-RW-1B, Falcon-7B~\citep{penedo2023refinedweb} \\
\midrule
\textbf{Training Datasets (3)} & News Category Dataset~\citep{misra2022news}, Wikipedia~\citep{wikidump}, CNN/DailyMail~\citep{hermann2015teaching} \\
\midrule
\textbf{Eval Transformers (7)} & GPT-2, Pythia-2.8B, Pythia-6.9B, OPT-2.7B, Llama-2-7B, Phi-2, Mistral-7B \\
\midrule
\textbf{Eval Non-Transformers (3)} & Mamba-2.8B (state-space), RWKV-3B (linear attention), RecurrentGemma-2B (gated recurrence) \\
\midrule
\textbf{Eval Datasets (4)} & AG News~\citep{zhang2015character}, WikiText-103~\citep{merity2016pointer}, XSum~\citep{narayan2018don}, Swallow-Code~\citep{fujii2025rewriting} \\
\bottomrule
\end{tabular}
\end{table}

\section{Feature Definitions}
\label{app:features}

Each input sample is a tensor of shape $(128, 154)$: 154 feature channels at each of 128 token positions. Features are extracted from both the fine-tuned target model and the pre-trained reference model.

\textit{Target model features (45 channels).} Per-token cross-entropy loss (1); logits of target's top-20 tokens (20); logits of target's bottom-20 tokens (20); logit assigned to ground-truth next token (1); log-normalized rank of ground-truth token (1); sequence-level mean and standard deviation of loss, broadcast to all positions (2).

\textit{Reference model features (45 channels).} Same structure as target features: per-token loss (1); reference's logits for the \emph{target's} top-20 and bottom-20 tokens (40); ground-truth logit and rank (2); global statistics (2).

\textit{Comparison features (64 channels).} Per-token loss difference $\ell_t^{\text{tgt}} - \ell_t^{\text{ref}}$ (1); sequence-level mean and standard deviation of loss difference (2); total log-likelihood ratio $\sum_t (\ell_t^{\text{tgt}} - \ell_t^{\text{ref}})$ broadcast to all positions (1); cross-model ranks for target's top-20 tokens in reference distribution (20); cross-model ranks for reference's top-20 tokens in target distribution (20); cross-model ranks for target's bottom-20 tokens (20).

\textit{Normalization.} Logit features are locally normalized by subtracting the maximum within each comparison group. Rank features are log-transformed and normalized by $\log(V+1)$ where $V$ is vocabulary size.

\section{Classifier Architecture Ablation}
\label{app:arch_ablation}

Table~\ref{tab:arch_ablation_full} presents full results for the classifier architecture comparison. All variants are trained on identical features from 30 model-dataset combinations (540,000 samples total); only the classifier architecture differs.

\begin{table}[h]
\caption{Classifier architecture ablation. The transformer processes features as a sequence; MLP and logistic regression flatten to a fixed-length vector; mean-pooled MLP averages features across positions before classification.}
\label{tab:arch_ablation_full}
\centering
\begin{tabular}{lcccc}
\toprule
Classifier & Train AUC & Eval AUC & Train TPR@1\% & Eval TPR@1\% \\
\midrule
Transformer (sequence) & 0.914 & \textbf{0.925} & 0.601 & \textbf{0.459} \\
MLP (flattened) & 0.885 & 0.877 & 0.483 & 0.343 \\
MLP (mean pooled) & 0.885 & 0.875 & 0.506 & 0.319 \\
Logistic Regression & 0.838 & 0.839 & 0.351 & 0.237 \\
\bottomrule
\end{tabular}
\end{table}

Sequence modeling contributes 5.0 AUC points over pooling (0.925 vs.\ 0.875).

\section{Diversity Ablation}
\label{app:diversity}

Table~\ref{tab:diversity_full} presents full results for the training diversity ablation. Total training samples are held constant at 18,000; only the number of model-dataset combinations varies.

\begin{table}[h]
\caption{Effect of training diversity on generalization with fixed total samples.}
\label{tab:diversity_full}
\centering
\begin{tabular}{cccccc}
\toprule
Combinations & Samples/Combo & Train AUC & Eval AUC & Train TPR@1\% & Eval TPR@1\% \\
\midrule
1 & 18,000 & 0.998 & 0.796 & 0.992 & 0.185 \\
10 & 1,800 & 0.875 & 0.843 & 0.371 & 0.244 \\
30 & 600 & 0.862 & \textbf{0.860} & 0.409 & \textbf{0.284} \\
\bottomrule
\end{tabular}
\end{table}

Training on a single combination achieves near-perfect training metrics (0.998 AUC) but generalizes poorly (0.796 AUC). With 30 combinations, the train-eval gap shrinks from 20.2 points to 0.2 points, demonstrating that diversity filters model-specific artifacts.

\section{Calibration Analysis}
\label{app:scatter}

\begin{table}[h]
\caption{Calibration comparison: LT-MIA vs.\ EZ-MIA on WikiText (where calibration gap is largest).}
\label{tab:scatter_data}
\centering
\small
\begin{tabular}{lcccc}
\toprule
Model & LT-MIA AUC & LT-MIA TPR@1\% & EZ-MIA AUC & EZ-MIA TPR@1\% \\
\midrule
GPT-2 & 0.980 & 0.632 & 0.971 & 0.058 \\
Pythia-2.8B & 0.944 & 0.492 & 0.867 & 0.016 \\
Pythia-6.9B & 0.796 & 0.137 & 0.911 & 0.036 \\
OPT-2.7B & 0.862 & 0.262 & 0.852 & 0.000 \\
Llama-2-7B & 0.781 & 0.130 & 0.733 & 0.020 \\
Phi-2 & 0.823 & 0.187 & 0.859 & 0.010 \\
Mistral-7B & 0.966 & 0.666 & 0.910 & 0.082 \\
Mamba-2.8B & 0.995 & 0.900 & 0.979 & 0.330 \\
RWKV-3B & 0.986 & 0.624 & 0.965 & 0.050 \\
RecurrentGemma-2B & 0.946 & 0.316 & 0.914 & 0.014 \\
\bottomrule
\end{tabular}
\end{table}

EZ-MIA's TPR@1\% collapses to near-zero on WikiText for most models despite achieving moderate-to-high AUC. LT-MIA maintains substantially higher TPR across all models, explaining why its advantage grows at stricter operating thresholds.

\end{document}